\documentclass[nohyperref]{article}

\usepackage{microtype}
\usepackage{subfigure}
\usepackage{booktabs} 

\usepackage{hyperref}

\usepackage[accepted]{icml2023}

\usepackage{amsmath}
\usepackage{amssymb}
\usepackage{mathtools}
\usepackage{amsthm}
\usepackage{bm}
\usepackage{wrapfig}
\usepackage{caption}

\usepackage{multirow}
\usepackage[capitalize,noabbrev]{cleveref}

\theoremstyle{plain}

\theoremstyle{definition}

\theoremstyle{remark}

\usepackage[textsize=tiny]{todonotes}

\icmltitlerunning{Ada-TTA: Towards Adaptive High-Quality Text-to-Talking Avatar Synthesis}

\begin{document}

\twocolumn[
\icmltitle{Ada-TTA: Towards Adaptive High-Quality Text-to-Talking Avatar Synthesis}




\icmlsetsymbol{equal}{*}
\icmlsetsymbol{interns}{\#}

\begin{icmlauthorlist}
\icmlauthor{Zhenhui Ye}{equal,interns,comp1,comp2}
\icmlauthor{Ziyue Jiang}{equal,interns,comp1,comp2}
\icmlauthor{Yi Ren}{equal,comp2}
\icmlauthor{Jinglin Liu}{comp2}
\icmlauthor{Chen Zhang}{interns,comp1,comp2}
\icmlauthor{Xiang Yin}{comp2}
\icmlauthor{Zejun Ma}{comp2}
\icmlauthor{Zhou Zhao}{comp1}
\end{icmlauthorlist}
\icmlaffiliation{comp1}{Zhejiang University}
\icmlaffiliation{comp2}{ByteDance}

\icmlcorrespondingauthor{Zhou Zhao}{zhaozhou@zju.edu.cn}


\vskip 0.3in
]



\printAffiliationsAndNotice{\icmlEqualContribution\textsuperscript{\#}Interns at ByteDance} 

\begin{abstract}
We are interested in a novel task, namely low-resource text-to-talking avatar. Given only a few-minute-long talking person video with the audio track as the training data and arbitrary texts as the driving input, we aim to synthesize high-quality talking portrait videos corresponding to the input text. This task has broad application prospects in the digital human industry but has not been technically achieved yet due to two challenges: (1) It is challenging to mimic the timbre from out-of-domain audio for a traditional multi-speaker Text-to-Speech system. (2) It is hard to render high-fidelity and lip-synchronized talking avatars with limited training data. In this paper, we introduce Adaptive Text-to-Talking Avatar (Ada-TTA), which (1) designs a generic zero-shot multi-speaker TTS model that well disentangles the text content, timbre, and prosody; and (2) embraces recent advances in neural rendering to achieve realistic audio-driven talking face video generation. With these designs, our method overcomes the aforementioned two challenges and achieves to generate identity-preserving speech and realistic talking person video. Experiments demonstrate that our method could synthesize realistic, identity-preserving, and audio-visual synchronized talking avatar videos.

\end{abstract}

\section{Introduction}
Recent years have witnessed an emergence of generative artificial intelligence in various domains, for instance, with a large language model (LLM)-based chatbot~\cite{adamopoulou2020chatbots}, we can obtain high-quality, natural, and realistic dialogue text content. Using an advanced text-to-speech (TTS) system~\cite{kim2021conditional, ren2021portaspeech, wang2023neural, ye2023clapspeech}, we can synthesize personalized speech given reference audio and plain texts. The development of neural rendering techniques also makes it possible to achieve high-fidelity and realistic talking face video generation (TFG)~\cite{wav2lip,guo2021ad,ye2023geneface} with only a few training samples. It is natural to think of combining the TTS model and TFG method so that the joint system could allow users to create a talking video with \textbf{only text input}. This joint system has broad potential applications such as news broadcasting, virtual lectures, and talking chatbots considering the recent advance of ChatGPT~\cite{ouyang2022training}.

However, previous TTS and TFG system typically requires a large amount of identity-specific data to produce satisfying personalized results\cite{ren2020fastspeech2, suwajanakorn2017synthesizing}, which raises challenges to real scenarios in which typically only a few-minute-long video of a target person is available. Motivated by this observation, we are interested in a novel task named low-resource text-to-talking avatar (TTA). Given only a few-minute-long talking person video with transcribed audio track as the training data, we aim to synthesize identity-preserving and audio-lip synchronized talking portrait videos given the driving input text. 

We first consider the challenges in TTS and TFG respectively. As for the TTS stage, the main challenge is how to properly preserve the timbre identity of the input audio~\cite{kharitonov2023speak}. A naive solution is to fine-tune a pre-trained TTS model on the given text-audio pairs. However, fine-tuning induces a large latency, and due to the limited amount of data, generalizability, the performance is not guaranteed; Another solution is to extract a speaker embedding of the input audio with an off-the-shelf toolkit~\cite{jia2018transfer}, which is known to introduce information loss and the identity-preserving quality is not satisfying. As for the TFG stage, the main challenge is to achieve high-fidelity and audio-visual synchronization given the limited amount of audio-video pairs. Some zero-shot methods~\cite{wav2lip,zhou2020makelttalk,PC-AVS} achieve good lip synchronization by training the model on big data, but the video quality is not as good. Some recent neural rendering-based methods~\cite{guo2021ad,tang2022real,liu2022semantic} achieve the goal of high fidelity, yet lip synchronization is poor due to the small amount of audio-lip training pairs.

In this paper, we propose Ada-TTA to handle the aforementioned problems. Ada-TTA is a joint system of TTS and TFG, which takes advantage of the most recent advances in each sub-task. To improve the identity-preserving power of the TTS model, we introduce a well-designed zero-shot multi-speaker TTS model trained on a 20,000-hour-long TTS dataset, which can synthesize high-quality personalized speech with only one short recording of an unseen speaker. To achieve high-fidelity and lip-synchronized talking face generation, we utilize the recently proposed GeneFace++~\cite{ye2023geneface++} as the TFG system, since it improves the lip-synchronization and system efficiency of the previous neural rendering-based methods while maintaining high fidelity. Combining the advantages of these two advanced systems in TTS and TFG, Ada-TTA achieves low-resource but high-quality text-to-talking avatar synthesis. The experiment shows good performance of our Ada-TTA in terms of synthesized speech and video. It also shows Ada-TTA outperforms the baseline from the perspective of objective and subjective metrics.

\section{Related works} \label{sec:rel}
Our work is majorly related to a low-resource personalized text-to-speech and talking face generation. We discuss related works from these two fields respectively.

Previous personalized speech generation approaches, also known as zero-shot multi-speaker TTS, can be categorized into speaker adaptation and speaker encoding methods. Traditional works~\cite{ren2019fastspeech,casanova2022yourtts, ye2022syntaspeech} are typically trained on small reading-style datasets and cannot generalize well for unseen speakers. Some recent works trained on large-scale multi-domain datasets demonstrate the effectiveness in zero-shot scenarios. Among them, some works~\cite{wang2023neural,shen2023naturalspeech} utilize the neural audio codec models to convert audio waveform into latent and consider it as the intermediate representation for speech generation, which ignores the intrinsic properties of speech attributes and may lead to inferior or uncontrollable results (e.g, degraded timbre identity similarity and uncontrollable prosody). By contrast, the proposed zero-shot multi-speaker TTS system disentangles the speech into different attributes and models each of them using architectures with appropriate inductive biases, which improves the identity-preserving power and prosody naturalness.

Traditional works in zero-shot/low-resource TFG~\cite{wav2lip, PC-AVS, zhou2020makelttalk} typically adopt a GAN-based renderer given a reference image of the identity, which fails to generate realistic and identity-preserving video. Recent works have embraced the neural radiance field (NeRF)~\cite{mildenhall2020nerf, guo2021ad} as it enables realistic 3D pose control and could achieve good video quality with limited data of an identity. Then Some works explore sample-efficient and time-efficient training~\cite{liu2022semantic,shen2022learning}. Recently, GeneFace++~\cite{ye2023geneface++}, a recently proposed method improves lip synchronization to OOD audio and achieves real-time inference, which promotes the NeRF-based TFG applicable to real-world scenarios.

\section{Proposed method} \label{sec:gdeq}

\begin{figure}[!t]
    \centering
    \includegraphics[width=1.0\linewidth]{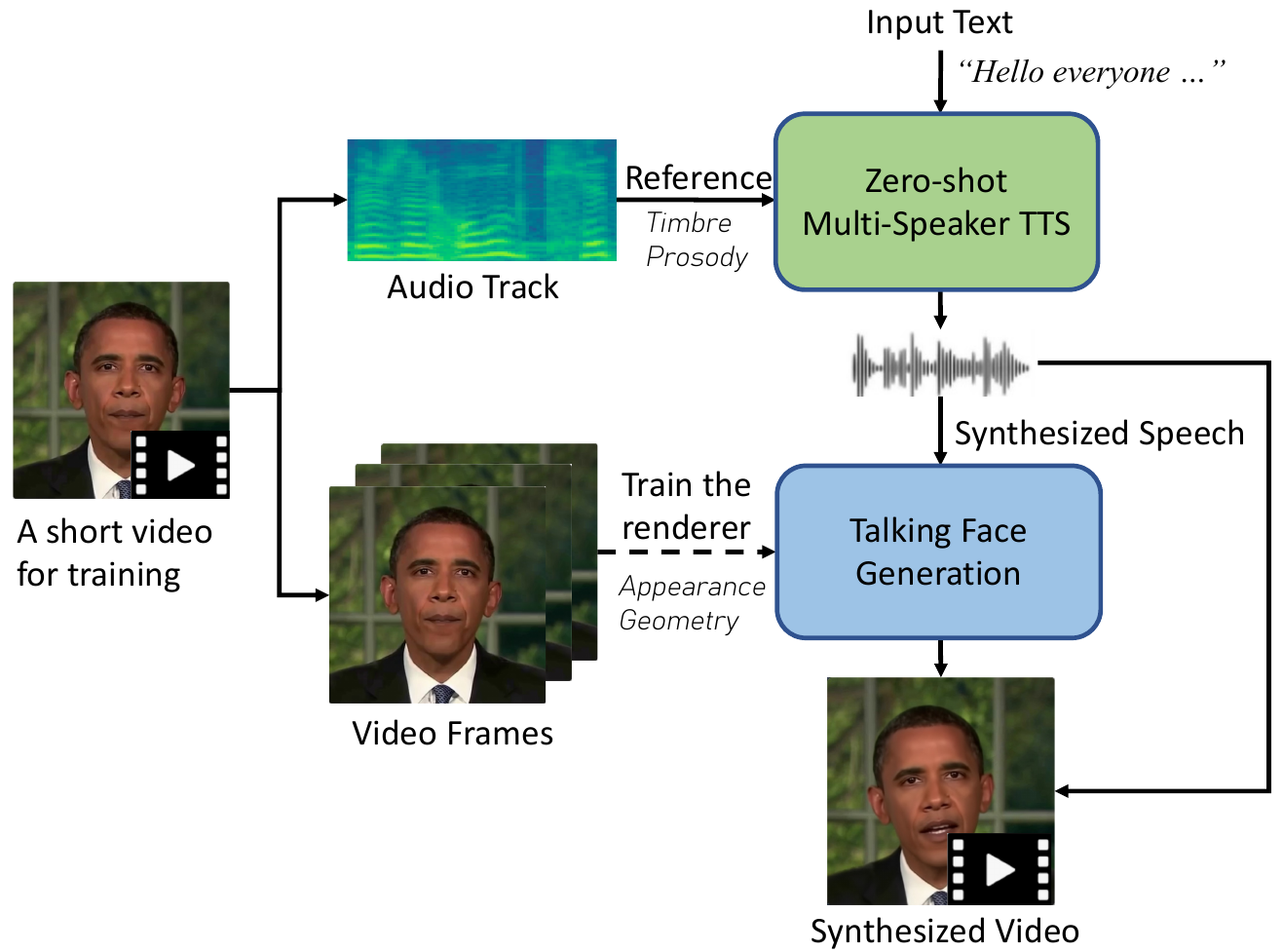}
    \caption{The overall pipeline of Ada-TTA. The dotted line denotes  that the process is only executed during the training phase.}
    \vspace{-3mm}
    \label{fig:overall_pipeline}
\end{figure}

The proposed system consists of two main modules: a zero-shot multi-speaker TTS module and a TFG module. First, the TTS model obtains a reference audio clip of the target identity and extracts the timbre and prosody pattern from it in an in-context-learning manner, then transforms the input text to the speech audio. Subsequently, the TFG module synthesizes the talking person portrait video synchronized with the input speech. The overall pipeline of our system is shown in Figure \ref{fig:overall_pipeline}.

\subsection{Zero-Shot Multi-Speaker TTS}
\begin{figure}[!t]
\centering
\includegraphics[width=0.8\linewidth]{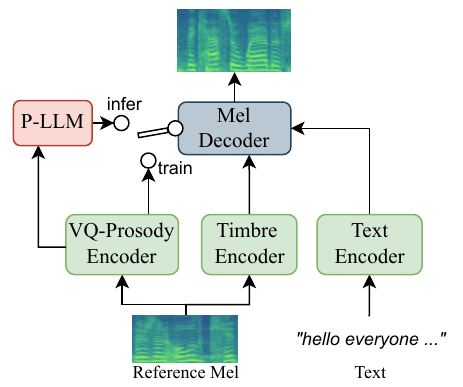}
\caption{The overall structure of our internal zero-shot multi-speaker TTS model.}
\vspace{-3mm}
\label{fig:mega}
\end{figure}

We use an internal zero-shot multi-speaker TTS model. As shown in Figure \ref{fig:mega}, it is a VQGAN-based~\cite{esser2021taming} TTS model which comprises a timbre encoder, a text encoder, a vector quantization (VQ) prosody encoder, and a mel decoder to disentangles the mel-spectrogram into speech attributes (e.g., prosody and timbre). We disentangle the mel-spectrogram with a carefully designed information bottleneck: 1) we use the text encoder to encode the phoneme sequence into the content representation; 2) we feed the reference mel-spectrogram sampled from a different speech of the same speaker to disentangle the timbre and content information and temporally average the output of the timbre encoder to get a one-dimensional global timbre vector. 3) we feed the first 20 bins in each mel-spectrogram frame into the prosody encoder, as it contains almost complete prosody and less timbre/content information. We also introduce a carefully-tuned vector quantization (VQ) layer and a phoneme-level downsampling layer to the prosody encoder to constrain the information flow. The correctly-designed bottleneck will learn to remove the content information and the global timbre information from the output of the prosody encoder, which ensures the performance of disentanglement. During training the GT mel-spectrogram is used to extract the prosody sequence. And during the inference phase, we need to predict the prosody sequence given the input text. To this end, leveraging the powerful in-context learning capability of LLMs, we learn a prosody large language model (P-LLM) that fits the distribution of prosody in an auto-regressive manner. To be specific, the P-LLM is a decoder-only transformer-based architecture that uses prosody codes from the reference speech as the prompt to generate the prosody codes for the target speech. Once the P-LLM is trained, during inference, we propose to use the content from the given text sequence, the timbre extracted from the prompt speech, and the prosody predicted by our P-LLM to generate the target speech for zero-shot personalized speech synthesis. 

\subsection{Talking Face Generation}

We adopt GeneFace++~\cite{ye2023geneface++} as the talking face generation model, which is a low-resource TFG method that could render lip-synchronized and high-fidelity talking face video in real-time. It compromises an audio-to-motion module to transform the input speech into facial landmarks, and a motion-to-video render to synthesize video frames given the landmark. We extract pitch contours and HuBERT features from the raw waveform as the audio representation, and select 68 facial landmarks from the reconstructed 3DMM mesh as the motion representation. The audio-to-motion module consists of a generic HuBERT-conditioned VAE and an identity-specific postnet to produce the audio-synchronized and identity-preserving facial landmark. The motion-to-video renderer is a landmark-conditioned NeRF that can freely control the facial expression and head pose by adjusting the facial landmark and the camera pose. To improve the system efficiency, the audio-to-motion module is built with a non-autoregressive network structure that processes the whole audio sequence in one forward, and the motion-to-video renderer adopts recent advances in Grid-based NeRF, which replaces previous computationally expensive dense MLP query with simple bi-linear indexing in a learnable grid. 

\subsection{Training and Inference}
The training procedure of the zero-shot multi-speaker TTS model has two stages. In the first stage, we train the VQGAN-based TTS model to reconstruct the mel-spectrogram given the input text, GT prosody code, and the reference mel-spectrogram. The training loss of the VQGAN-based TTS is as follows:
\begin{equation}
    \mathcal{L} = ||y - \hat{y}||_2^2 + \mathcal{L}_{VQ} + \mathcal{L}_{Adv}
\end{equation}
where $||y - \hat{y}||_2^2$ is the L2 loss on mel-spectrogram, $\mathcal{L}_{VQ}$ is the VQVAE loss function~\cite{van2017neural,esser2021taming} and $\mathcal{L}_{Adv}$ is the LSGAN-styled adversarial loss~\cite{mao2017least}. Then in the second stage, we train the P-LLM to predict the prosody code given the input text and previous prosody code sequence. The P-LLM is trained in a teacher-forcing mode in the training stage via the cross-entropy loss.

In training our talking face generation model, we use the pre-trained audio-to-motion module in GeneFace++ and train the NeRF-based renderer from scratch. To further improve the image quality, we adopt VGG perceptual loss on the lip part of the predicted image.

During inference, the zero-shot multi-speaker TTS model synthesizes identity-preserving speech given the input text, then the talking face generation model takes the synthesized audio as input and generates the talking portrait images. Finally, the synthesized audio and video frames are integrated into the final video.

\section{Experiments} 

\paragraph{Training Details.} As for the TTS model, to improve the timbre and prosody generalizability, we scale up the TTS model to 222.5M parameters and adopt GigaSpeech~\cite{chen2021gigaspeech}, a 10,000-hour-long English TTS dataset for training the model. We train this large-scale TTS model on 8 NVIDIA A100 GPUs with a batch size of 30 sentences on each GPU. It takes 420k steps for convergence and no further fine-tuning is needed. As for the talking face generation, we use the pre-trained audio-to-motion module provided by GeneFace++~\cite{ye2023geneface++} and train the NeRF-based renderer for each specific identity for 320k steps, which takes about 6 hours on 1 NVIDIA A100 GPU. The talking person videos used to train the renderer are about 3 minutes in length.

\paragraph{Comparison baseline.} Since there is no publicly available low-resource text-to-avatar system, we construct a baseline by combining YourTTS~\cite{casanova2022yourtts}, a recently proposed zero-shot multi-speaker TTS, and Wav2Lip~\cite{wav2lip}, a state-of-the-art few-shot talking face generation method from the perspective of lip synchronization. We denote the baseline as \textit{YourTTS+Wav2Lip}. Note that during inference Wav2Lip takes the whole training video as input and only regenerates the lip part. By contrast, our model renders the whole frame.

\paragraph{Evaluation Metrics.} We conduct the CMOS (comparative mean opinion score) test to evaluate the performance of the text-to-avatar systems. We analyze the CMOS in three aspects: CMOS-A (Only analyze the audio, including speaker similarity, prosody, and audio quality), CMOS-V (Only analyze the video, including image fidelity, 3D realness, and identity preserving), CMOS (the overall opinion score of the synthesized video). We tell the tester to focus on one corresponding aspect and ignore the other aspect when scoring. As for the subjective evaluation, we use the WavLM~\cite{chen2022wavlm} model fine-tuned for speaker verification to compute the cosine similarity score between the ground-truth speech and synthesized speech; we use FID to measure the image quality of the synthesized video.

\begin{figure}[!t]
    \centering
    \includegraphics[width=1.0\linewidth]{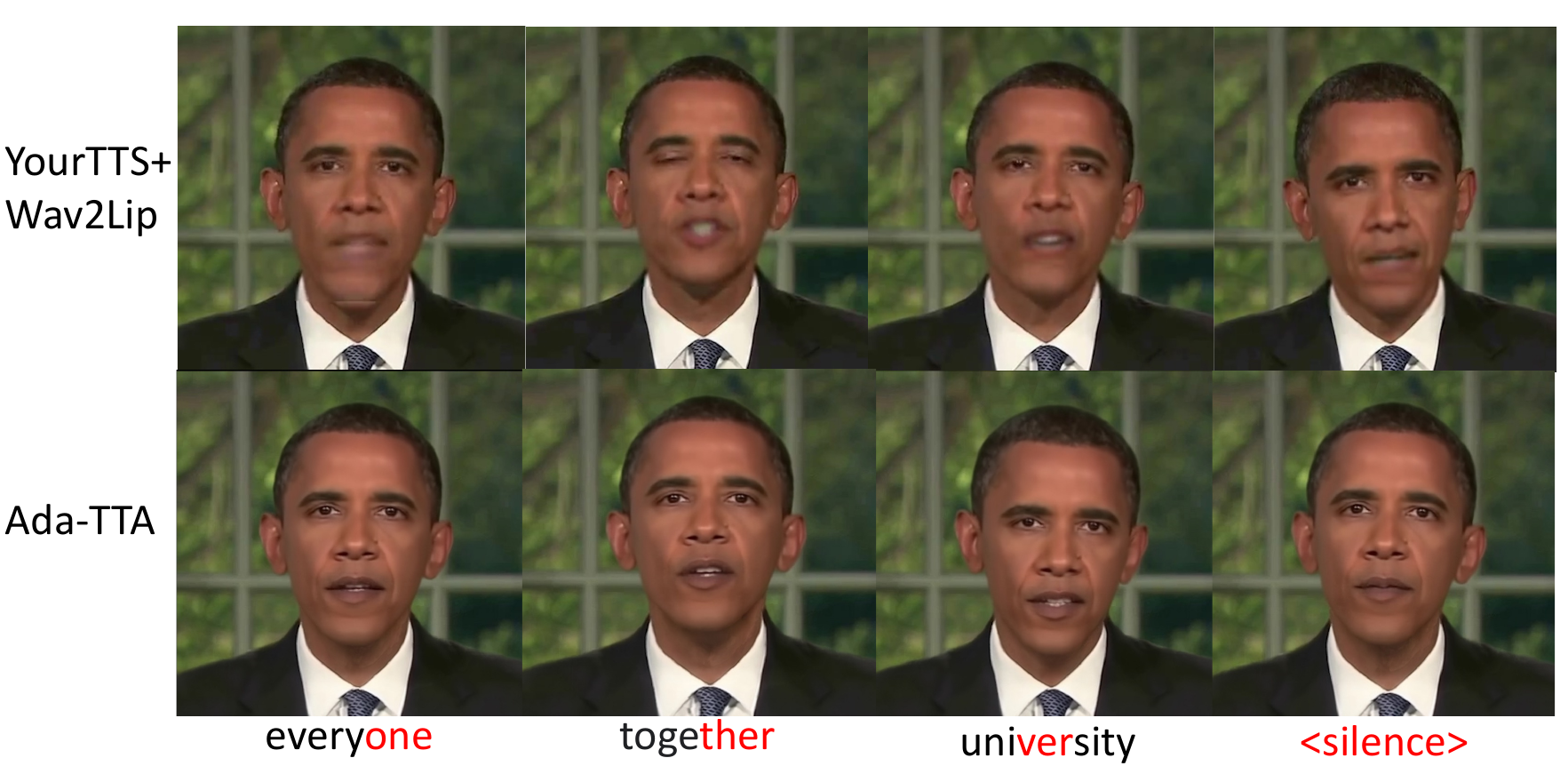}
    \caption{The video frames generated by the TTA systems.}
    \vspace{-3mm}
    \label{fig:demo_imgs}
\end{figure}

\paragraph{Experiment Results.} 
As shown in Table \ref{tab:subjective}, our method achieves a higher speaker similarity than YourTTS, proving the effectiveness of our zero-shot multi-speaker TTS model to preserve the identity of the input reference audio. We also notice our method achieves better FID than Wav2Lip, which denotes a better image quality of the rendered video frames. We further perform the CMOS test to evaluate the performance in terms of human perception. The results are listed in Table \ref{tab:cmos}. We find that prefer the synthesized videos by our method in terms of audio quality (CMOS-A), video quality (CMOS-V), and overall quality (CMOS). To make a better qualitative comparison, we recommend the reader to watch a demo video \footnote{The link for demo video is \url{https://genefaceplusplus.github.io/GeneFace++/ada_tta.mp4}}. We also provide some keyframes in Figure \ref{fig:demo_imgs} to compare the performance of our Ada-TTA and the baseline. We can observe that compared with the baseline, Ada-TTA produces (1) better speech in terms of high timbre similarity and good audio quality; and (2) better video with high lip-synchronization, good image fidelity, and free head pose control. 

 \begin{table}[t]
    \caption{Objective evaluation of the TTA systems. Spk-Sim denotes cosine speaker similarity. }
    \begin{center}
        \centering
            \begin{tabular}{lcc}
                \toprule
                \textbf{Method} & Spk-Sim$\uparrow$ & FID$\downarrow$ \\
                \midrule
                {YourTTS + Wav2Lip} &  0.9392 & 55.43  \\
                {Ada-TTA (Ours)} &  \textbf{0.9854} & \textbf{28.36}  \\
                \bottomrule
            \end{tabular}
    \end{center}
    \label{tab:subjective}
\end{table}

\begin{table}[t]
    \caption{CMOS Results of the TTA systems. The error bar is standard deviation. Y+W denotes YourTTS+Wav2Lip.}
    \begin{center}
        \centering
        \small
            \begin{tabular}{lccc}
                \toprule
                \textbf{Method} & CMOS-A & CMOS-V &  CMOS \\
                \midrule
                {Y+W} &  0.00 & 0.00 &0.00 \\
                {Ada-TTA} & $+0.84\pm 0.50$ & $+0.76\pm 0.42$ & $+0.74\pm 0.31$ \\
                \bottomrule
            \end{tabular}
    \end{center}
    \label{tab:cmos}
\end{table}

\section{Conclusions}

In this paper, we present Ada-TTA, an adaptive high-quality text-to-talking avatar synthesis system. Given only a few-minute-long talking person video as training data, with Ada-TTA we can synthesize identity-preserving speech given arbitrary input text, and generate the lip-synchronized video. We describe a zero-shot multi-speaker TTS model and a high-quality talking face generation method used to construct the Ada-TTA system. Our experiments demonstrate our system's ability to synthesize realistic speech and video at a limited data scale.

\bibliography{example_paper}

\begin{thebibliography}{29}
\providecommand{\natexlab}[1]{#1}
\providecommand{\url}[1]{\texttt{#1}}
\expandafter\ifx\csname urlstyle\endcsname\relax
  \providecommand{\doi}[1]{doi: #1}\else
  \providecommand{\doi}{doi: \begingroup \urlstyle{rm}\Url}\fi

\bibitem[Adamopoulou \& Moussiades(2020)Adamopoulou and
  Moussiades]{adamopoulou2020chatbots}
Adamopoulou, E. and Moussiades, L.
\newblock Chatbots: History, technology, and applications.
\newblock \emph{Machine Learning with Applications}, 2:\penalty0 100006, 2020.

\bibitem[Casanova et~al.(2022)Casanova, Weber, Shulby, Junior, G{\"o}lge, and
  Ponti]{casanova2022yourtts}
Casanova, E., Weber, J., Shulby, C.~D., Junior, A.~C., G{\"o}lge, E., and
  Ponti, M.~A.
\newblock Yourtts: Towards zero-shot multi-speaker tts and zero-shot voice
  conversion for everyone.
\newblock In \emph{International Conference on Machine Learning}, pp.\
  2709--2720. PMLR, 2022.

\bibitem[Chen et~al.(2021)Chen, Chai, Wang, Du, Zhang, Weng, Su, Povey, Trmal,
  Zhang, et~al.]{chen2021gigaspeech}
Chen, G., Chai, S., Wang, G., Du, J., Zhang, W.-Q., Weng, C., Su, D., Povey,
  D., Trmal, J., Zhang, J., et~al.
\newblock Gigaspeech: An evolving, multi-domain asr corpus with 10,000 hours of
  transcribed audio.
\newblock \emph{arXiv preprint arXiv:2106.06909}, 2021.

\bibitem[Chen et~al.(2022)Chen, Wang, Chen, Wu, Liu, Chen, Li, Kanda, Yoshioka,
  Xiao, et~al.]{chen2022wavlm}
Chen, S., Wang, C., Chen, Z., Wu, Y., Liu, S., Chen, Z., Li, J., Kanda, N.,
  Yoshioka, T., Xiao, X., et~al.
\newblock Wavlm: Large-scale self-supervised pre-training for full stack speech
  processing.
\newblock \emph{IEEE Journal of Selected Topics in Signal Processing},
  16\penalty0 (6):\penalty0 1505--1518, 2022.

\bibitem[Esser et~al.(2021)Esser, Rombach, and Ommer]{esser2021taming}
Esser, P., Rombach, R., and Ommer, B.
\newblock Taming transformers for high-resolution image synthesis.
\newblock In \emph{Proceedings of the IEEE/CVF conference on computer vision
  and pattern recognition}, pp.\  12873--12883, 2021.

\bibitem[Guo et~al.(2021)Guo, Chen, Liang, Liu, Bao, and Zhang]{guo2021ad}
Guo, Y., Chen, K., Liang, S., Liu, Y.-J., Bao, H., and Zhang, J.
\newblock Ad-nerf: Audio driven neural radiance fields for talking head
  synthesis.
\newblock In \emph{ICCV}, pp.\  5784--5794, 2021.

\bibitem[Jia et~al.(2018)Jia, Zhang, Weiss, Wang, Shen, Ren, Nguyen, Pang,
  Lopez~Moreno, Wu, et~al.]{jia2018transfer}
Jia, Y., Zhang, Y., Weiss, R., Wang, Q., Shen, J., Ren, F., Nguyen, P., Pang,
  R., Lopez~Moreno, I., Wu, Y., et~al.
\newblock Transfer learning from speaker verification to multispeaker
  text-to-speech synthesis.
\newblock \emph{Advances in neural information processing systems}, 31, 2018.

\bibitem[Kharitonov et~al.(2023)Kharitonov, Vincent, Borsos, Marinier, Girgin,
  Pietquin, Sharifi, Tagliasacchi, and Zeghidour]{kharitonov2023speak}
Kharitonov, E., Vincent, D., Borsos, Z., Marinier, R., Girgin, S., Pietquin,
  O., Sharifi, M., Tagliasacchi, M., and Zeghidour, N.
\newblock Speak, read and prompt: High-fidelity text-to-speech with minimal
  supervision.
\newblock \emph{arXiv preprint arXiv:2302.03540}, 2023.

\bibitem[Kim et~al.(2021)Kim, Kong, and Son]{kim2021conditional}
Kim, J., Kong, J., and Son, J.
\newblock Conditional variational autoencoder with adversarial learning for
  end-to-end text-to-speech.
\newblock In \emph{ICML}. PMLR, 2021.

\bibitem[Liu et~al.(2022)Liu, Xu, Wu, Zhou, Wu, and Zhou]{liu2022semantic}
Liu, X., Xu, Y., Wu, Q., Zhou, H., Wu, W., and Zhou, B.
\newblock Semantic-aware implicit neural audio-driven video portrait
  generation.
\newblock \emph{arXiv preprint arXiv:2201.07786}, 2022.

\bibitem[Mao et~al.(2017)Mao, Li, Xie, Lau, Wang, and
  Paul~Smolley]{mao2017least}
Mao, X., Li, Q., Xie, H., Lau, R.~Y., Wang, Z., and Paul~Smolley, S.
\newblock Least squares generative adversarial networks.
\newblock In \emph{ICCV}, 2017.

\bibitem[Mildenhall et~al.(2020)Mildenhall, Srinivasan, Tancik, Barron,
  Ramamoorthi, and Ng]{mildenhall2020nerf}
Mildenhall, B., Srinivasan, P.~P., Tancik, M., Barron, J.~T., Ramamoorthi, R.,
  and Ng, R.
\newblock Nerf: Representing scenes as neural radiance fields for view
  synthesis.
\newblock In \emph{ECCV}, pp.\  405--421. Springer, 2020.

\bibitem[Ouyang et~al.(2022)Ouyang, Wu, Jiang, Almeida, Wainwright, Mishkin,
  Zhang, Agarwal, Slama, Ray, et~al.]{ouyang2022training}
Ouyang, L., Wu, J., Jiang, X., Almeida, D., Wainwright, C., Mishkin, P., Zhang,
  C., Agarwal, S., Slama, K., Ray, A., et~al.
\newblock Training language models to follow instructions with human feedback.
\newblock \emph{NeurIPS}, 2022.

\bibitem[Prajwal et~al.(2020)Prajwal, Mukhopadhyay, Namboodiri, and
  Jawahar]{wav2lip}
Prajwal, K., Mukhopadhyay, R., Namboodiri, V.~P., and Jawahar, C.
\newblock A lip sync expert is all you need for speech to lip generation in the
  wild.
\newblock In \emph{ACM MM}, pp.\  484--492, 2020.

\bibitem[Ren et~al.(2019)Ren, Ruan, Tan, Qin, Zhao, Zhao, and
  Liu]{ren2019fastspeech}
Ren, Y., Ruan, Y., Tan, X., Qin, T., Zhao, S., Zhao, Z., and Liu, T.-Y.
\newblock Fastspeech: Fast, robust and controllable text to speech.
\newblock \emph{Advances in neural information processing systems}, 32, 2019.

\bibitem[Ren et~al.(2020)Ren, Hu, Tan, Qin, Zhao, Zhao, and
  Liu]{ren2020fastspeech2}
Ren, Y., Hu, C., Tan, X., Qin, T., Zhao, S., Zhao, Z., and Liu, T.-Y.
\newblock Fastspeech 2: Fast and high-quality end-to-end text to speech.
\newblock \emph{arXiv preprint arXiv:2006.04558}, 2020.

\bibitem[Ren et~al.(2021)Ren, Liu, and Zhao]{ren2021portaspeech}
Ren, Y., Liu, J., and Zhao, Z.
\newblock Portaspeech: Portable and high-quality generative text-to-speech.
\newblock \emph{NIPS}, 34:\penalty0 13963--13974, 2021.

\bibitem[Shen et~al.(2023)Shen, Ju, Tan, Liu, Leng, He, Qin, Zhao, and
  Bian]{shen2023naturalspeech}
Shen, K., Ju, Z., Tan, X., Liu, Y., Leng, Y., He, L., Qin, T., Zhao, S., and
  Bian, J.
\newblock Naturalspeech 2: Latent diffusion models are natural and zero-shot
  speech and singing synthesizers.
\newblock \emph{arXiv preprint arXiv:2304.09116}, 2023.

\bibitem[Shen et~al.(2022)Shen, Li, Zhu, Duan, Zhou, and Lu]{shen2022learning}
Shen, S., Li, W., Zhu, Z., Duan, Y., Zhou, J., and Lu, J.
\newblock Learning dynamic facial radiance fields for few-shot talking head
  synthesis.
\newblock In \emph{ECCV}, 2022.

\bibitem[Suwajanakorn et~al.(2017)Suwajanakorn, Seitz, and
  Kemelmacher-Shlizerman]{suwajanakorn2017synthesizing}
Suwajanakorn, S., Seitz, S.~M., and Kemelmacher-Shlizerman, I.
\newblock Synthesizing obama: learning lip sync from audio.
\newblock \emph{ACM Transactions on Graphics (ToG)}, 36\penalty0 (4):\penalty0
  1--13, 2017.

\bibitem[Tang et~al.(2022)Tang, Wang, Zhou, Chen, He, Hu, Liu, Zeng, and
  Wang]{tang2022real}
Tang, J., Wang, K., Zhou, H., Chen, X., He, D., Hu, T., Liu, J., Zeng, G., and
  Wang, J.
\newblock Real-time neural radiance talking portrait synthesis via
  audio-spatial decomposition.
\newblock \emph{arXiv preprint arXiv:2211.12368}, 2022.

\bibitem[Van Den~Oord et~al.(2017)Van Den~Oord, Vinyals, et~al.]{van2017neural}
Van Den~Oord, A., Vinyals, O., et~al.
\newblock Neural discrete representation learning.
\newblock \emph{Advances in neural information processing systems}, 30, 2017.

\bibitem[Wang et~al.(2023)Wang, Chen, Wu, Zhang, Zhou, Liu, Chen, Liu, Wang,
  Li, et~al.]{wang2023neural}
Wang, C., Chen, S., Wu, Y., Zhang, Z., Zhou, L., Liu, S., Chen, Z., Liu, Y.,
  Wang, H., Li, J., et~al.
\newblock Neural codec language models are zero-shot text to speech
  synthesizers.
\newblock \emph{arXiv preprint arXiv:2301.02111}, 2023.

\bibitem[Ye et~al.(2022)Ye, Zhao, Ren, and Wu]{ye2022syntaspeech}
Ye, Z., Zhao, Z., Ren, Y., and Wu, F.
\newblock Syntaspeech: syntax-aware generative adversarial text-to-speech.
\newblock \emph{arXiv preprint arXiv:2204.11792}, 2022.

\bibitem[Ye et~al.(2023{\natexlab{a}})Ye, He, Jiang, Huang, Huang, Liu, Ren,
  Yin, Ma, and Zhao]{ye2023geneface++}
Ye, Z., He, J., Jiang, Z., Huang, R., Huang, J., Liu, J., Ren, Y., Yin, X., Ma,
  Z., and Zhao, Z.
\newblock Geneface++: Generalized and stable real-time audio-driven 3d talking
  face generation.
\newblock \emph{arXiv preprint arXiv:2305.00787}, 2023{\natexlab{a}}.

\bibitem[Ye et~al.(2023{\natexlab{b}})Ye, Huang, Ren, Jiang, Liu, He, Yin, and
  Zhao]{ye2023clapspeech}
Ye, Z., Huang, R., Ren, Y., Jiang, Z., Liu, J., He, J., Yin, X., and Zhao, Z.
\newblock Clapspeech: Learning prosody from text context with contrastive
  language-audio pre-training.
\newblock \emph{arXiv preprint arXiv:2305.10763}, 2023{\natexlab{b}}.

\bibitem[Ye et~al.(2023{\natexlab{c}})Ye, Jiang, Ren, Liu, He, and
  Zhao]{ye2023geneface}
Ye, Z., Jiang, Z., Ren, Y., Liu, J., He, J., and Zhao, Z.
\newblock Geneface: Generalized and high-fidelity audio-driven 3d talking face
  synthesis.
\newblock In \emph{ICLR}, 2023{\natexlab{c}}.

\bibitem[Zhou et~al.(2021)Zhou, Sun, Wu, Loy, Wang, and Liu]{PC-AVS}
Zhou, H., Sun, Y., Wu, W., Loy, C.~C., Wang, X., and Liu, Z.
\newblock Pose-controllable talking face generation by implicitly modularized
  audio-visual representation.
\newblock In \emph{CVPR}, pp.\  4176--4186, 2021.

\bibitem[Zhou et~al.(2020)Zhou, Han, Shechtman, Echevarria, Kalogerakis, and
  Li]{zhou2020makelttalk}
Zhou, Y., Han, X., Shechtman, E., Echevarria, J., Kalogerakis, E., and Li, D.
\newblock Makelttalk: speaker-aware talking-head animation.
\newblock \emph{ACM Transactions on Graphics (TOG)}, 39\penalty0 (6):\penalty0
  1--15, 2020.

\end{thebibliography}
\bibliographystyle{icml2023}

\end{document}